\documentclass{article}

\usepackage{microtype}
\usepackage{graphicx}
\usepackage{subfigure}
\usepackage{booktabs} 
\usepackage{multirow}
\usepackage{adjustbox}
\usepackage[table]{xcolor}
\usepackage{hyperref}

\usepackage[accepted]{icml2024}

\usepackage{amsmath}
\usepackage{amssymb}
\usepackage{mathtools}
\usepackage{amsthm}

\usepackage[capitalize,noabbrev]{cleveref}

\theoremstyle{plain}

\theoremstyle{definition}

\theoremstyle{remark}

\usepackage[textsize=tiny]{todonotes}

\begin{document}

\twocolumn[
\icmltitle{InfiMM-HD: A Leap Forward in High-Resolution Multimodal Understanding}



\icmlsetsymbol{equal}{*}

\begin{icmlauthorlist}
\icmlauthor{Haogeng Liu}{equal,yyy,comp}
\icmlauthor{Quanzeng You}{equal,sch}
\icmlauthor{Xiaotian Han}{sch}
\icmlauthor{Yiqi Wang}{sch}
\icmlauthor{Bohan Zhai}{sch}
\icmlauthor{Yongfei Liu}{sch}
\icmlauthor{Yunzhe Tao}{sch}
\icmlauthor{Huaibo Huang}{yyy,comp}
\icmlauthor{Ran He}{equal,yyy,comp}
\icmlauthor{Hongxia Yang}{equal,sch}
\end{icmlauthorlist}

\icmlaffiliation{yyy}{MAIS \& CRIPAC, Institute of Automation, Chinese Academy of Sciences, China.}
\icmlaffiliation{comp}{School of Artificial Intelligence, University of Chinese Academy of Sciences, Beijing, China.}
\icmlaffiliation{sch}{ByteDance, Inc.}

\icmlcorrespondingauthor{Haogeng Liu}{liuhaogeng22@mails.ucas.ac.cn}

\icmlkeywords{Machine Learning, ICML, Multimodal Large Language Models}

\vskip 0.3in
]




\printAffiliationsAndNotice{\icmlEqualContribution} 

\begin{abstract}

Multimodal Large Language Models (MLLMs) have experienced significant advancements recently.
Nevertheless, challenges persist in the accurate recognition and comprehension of intricate details within high-resolution images. 
Despite being indispensable for the development of robust MLLMs, this area remains underinvestigated. 
To tackle this challenge, our work introduces InfiMM-HD, a novel architecture specifically designed for processing images of different resolutions with low computational overhead.
This innovation facilitates the enlargement of MLLMs to higher-resolution capabilities. 
InfiMM-HD incorporates a cross-attention module and visual windows to reduce computation costs. 
By integrating this architectural design with a four-stage training pipeline, our model attains improved visual perception efficiently and cost-effectively.
Empirical study underscores the robustness and effectiveness of InfiMM-HD, opening new avenues for exploration in related areas. Codes and models can be found {\href{https://huggingface.co/Infi-MM/infimm-hd}{https://huggingface.co/Infi-MM/infimm-hd}}

\end{abstract}

\section{Introduction}

\begin{figure}[ht]
    \centering
    \includegraphics[width=8cm]{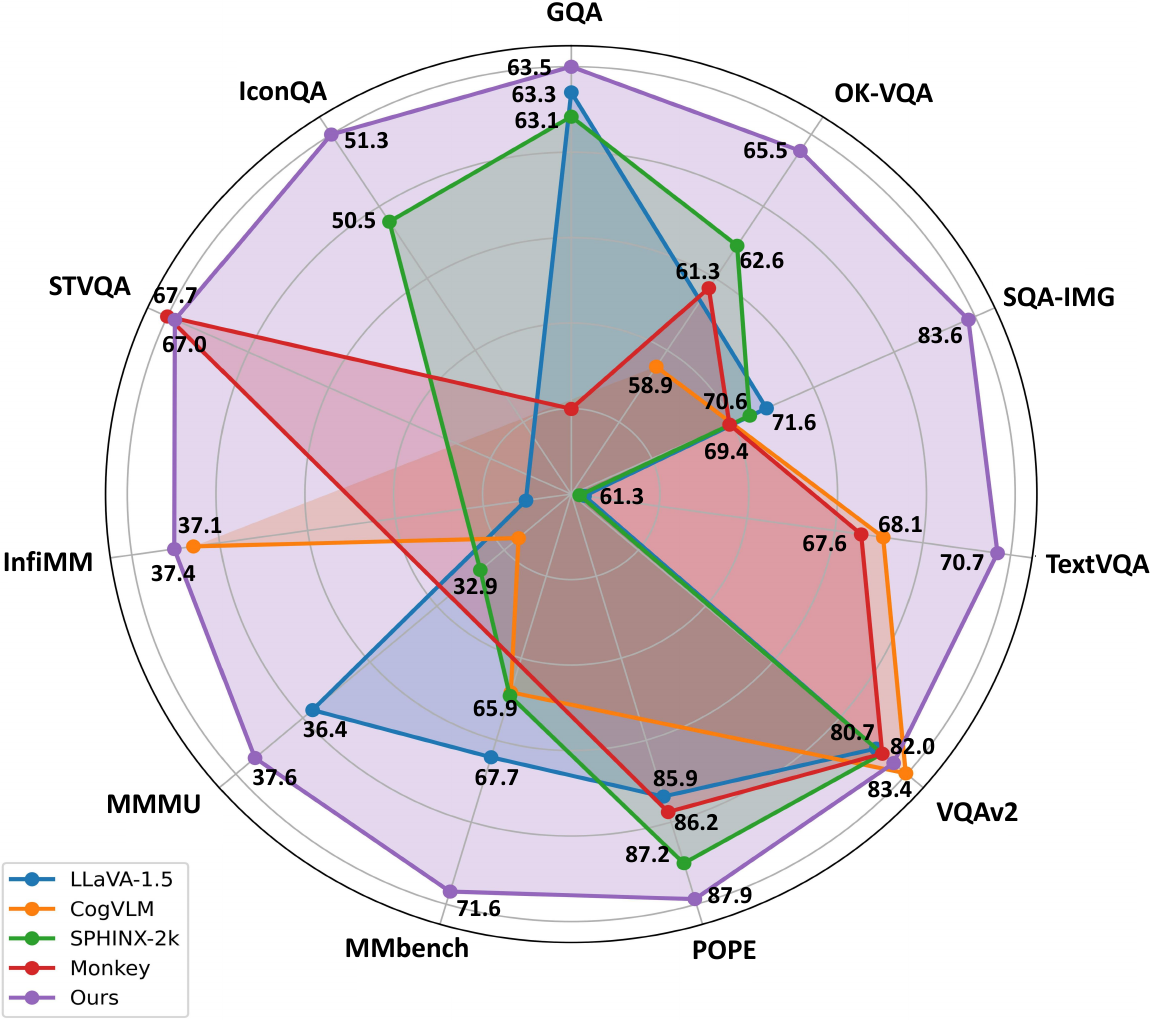}
    \caption{InfiMM-HD showcases superior performance across various tasks, thanks to its enhanced training pipeline and high-resolution inputs, consistently outperforming recent methods in downstream tasks.}
    \label{fig:enter-label}
\end{figure}

\begin{figure*}[ht]
    \label{fig:demo_infimm}
    \centering
    \includegraphics[width=16cm]{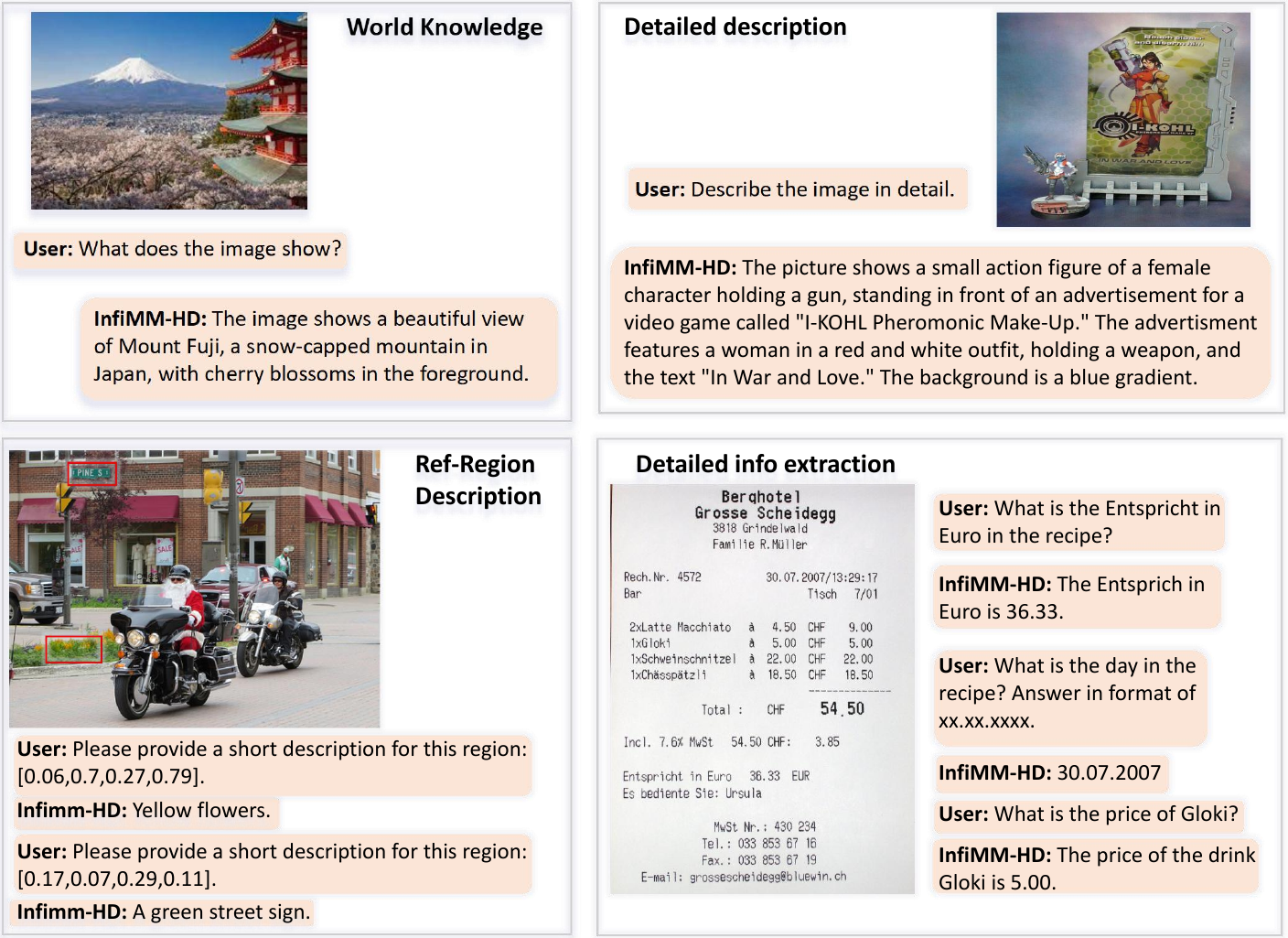}
    \caption{Example outputs by InfiMM-HD, highlighting the model's adeptness in fine-grained visual perception.}
\end{figure*}

The landscape of Multimodal Large Language Models (MLLMs) has been revolutionized by integrating pretrained vision encoders with Large Language Models (LLMs) \cite{2023arXiv231111567H, wang2024exploring, han2024coco}, a trend exemplified by developments in Flamingo \cite{alayrac2022flamingo}, BLIP-2 \cite{li2023blip2}, LLaVA \cite{liu2023visual}, MiniGPT-4 \cite{zhu2023minigpt4} and etc. 
MLLMs can exhibit emergent vision-language capabilities~\cite{yang2023dawn}. 
For example, they can write codes according to images, convert plots in images into Markdown format tables, and perform web browsing, by leveraging combined single-modal capabilities from pretrained vision encoders and LLMs. 

Effective integration of pretrained vision encoders with Large Language Models in MLLMs relies on carefully designed vision-language bridging modules.
These modules address two critical aspects: the transformation and alignment of visual tokens to LLM-compatible formats, and the subsequent utilization of these transformed tokens.

For the transformation and alignment of visual tokens, models such as Flamingo~\cite{alayrac2022flamingo} and BLIP-2~\cite{li2023blip2} employ Perceiver-Resampler/Q-Former techniques for transforming visual tokens, offering flexibility and nuanced alignment with language counterparts but at the cost of higher computational demands and potential information loss~\cite{cha2023honeybee}. 
Conversely, models such as LLaVA and MiniGPT-v2 take a different route, using simpler Multi-Layer Perceptron (MLP) approaches.
While this reduces computational complexity and the number of learnable parameters, it might not capture the full intricacies of visual data.

The integration phase of visual and language tokens is equally vital. 
Flamingo-style architectures employ cross-attention mechanisms, facilitating intricate interactions between token types without expanding the token sequence length. 
This approach efficiently manages the computational load. 
LLaVA-style models, however, use a direct concatenation method, which, while straightforward, increases token sequence length and computational complexity.

Though different, both architecture utilized low image resolution due to pretrained Vision Transformer (ViT) encoders~\cite{jiang2023clip, radford2021learning, he2021masked}. 
Low resolution suffices for basic image-level semantic understanding but falls short for detailed, region-level analysis. 
Recent efforts~\cite{wang2023cogvlm, li2023monkey, lin2023sphinx} aim to enable MLLMs to handle higher-resolution images.
However, significant challenges remain, primarily because the computational demands tend to increase quadratically in relation to the sequence length for larger images.
For instance, increasing the image resolution from $224\times224$~\cite{dosovitskiy2021image} to $448\times448$ multiplies the self-attention computation  by 16 times.

To address these challenges, we introduce InfiMM-HD, an innovative MLLM architecture designed for processing high-resolution image. 
InfiMM-HD innovatively merges methodologies from both Flamingo and LLaVA styles in MLLMs. 
For the transformation and alignment phase, it adopts an MLP-based approach, akin to LLaVA, effectively transforming and aligning visual tokens into formats compatible with LLMs. This strategy balances computational efficiency with precise processing. 
In the integration phase, InfiMM-HD utilizes a cross-attention mechanism, reminiscent of Flamingo-style MLLMs, to seamlessly incorporate visual token features with language tokens. 
This approach mitigates the computational challenges associated with longer token sequences as previously mentioned. 
Notably, while the exploration of high-resolution image input capabilities in Flamingo-style MLLMs is still an emerging area of research, InfiMM-HD marks a significant, pioneering advancement in this domain, blending the best of both worlds to enhance MLLM performance with high-resolution visual inputs.

To overcome the resolution constraints of pretrained vision encoders, InfiMM-HD is strategically trained in four stages, enhancing resolution handling while maintaining vision-language alignment. 
Initially, the model is pretrained with $224\times224$ resolution images for efficient visual-language alignment. 
Subsequently, it continues pretraining with interpolated positional embeddings for $448\times448$ images from multiple datasets, keeping the LLMs frozen.
This is followed by training with full-resolution images, resized to the nearest multiple of $448\times448$, added with 2D positional embeddings and crop to multiple subimages.
In the final stage, the model undergoes visual instruction fine-tuning, freezing the vision encoder and making LLM trainable to enhance instruction-following capability.
This structured training approach is crucial for the model's adaptability and performance across various input resolutions.
The contributions of our work can be summarized as follows:
\begin{itemize}
    \item We present InfiMM-HD, a pioneering MLLM that employs an MLP-based approach for visual token transformation and alignment, coupled with a Flamingo-style cross-attention mechanism for enhanced and efficient integration of transformed visual and language tokens. It is uniquely designed to seamlessly process \textbf{high-resolution} image inputs.
    \item We present a four-stage training pipeline that effectively achieves a high-resolution Multimodal Large Language Model with reduced training cost, from initial low-resolution pretraining stage, to continue pretraining stage for knowledge injection and alignment, to dynamic resolution adaption stage for high resolution adoption and finally go through visual instruction fine-tuning stage.
    \item 
    Experiments conducted across diverse benchmarks showcase the remarkable proficiency of our model in the realm of visual perception. Additionally, comprehensive ablation studies underscore the distinctive superiority of our design within the context of cross-attention-style Multimodal Language Model architectures.
\end{itemize}

\section{Related Work}
The advent of Large Language Models (LLMs) has catalyzed the development of MLLMs. 
Flamingo \cite{alayrac2022flamingo} integrates pretrained language models into the MLLM paradigm, employing a gated-cross attention mechanism to fuse visual information into textual sequences. 
In contrast, BLIP-2 \cite{li2023blip2}, MiniGPT4 \cite{zhu2023minigpt4}, and LLaVA \cite{liu2023visual} propose a paradigm shift, transforming visual signals into soft tokens and directly incorporating them into language models. 
Shikra \cite{chen2023shikra} concentrates on referential dialogue. OtterHD \cite{li2023otterhd} fine-tunes Fuyu-8B \cite{fuyu-8b} with instructional guidance, enabling ViT-free MLLMs.

Despite the progress we have seen, some problems still exist. 
\cite{zhai2023halleswitch} points out that misalignment between visual representation and language causes hallucination. 
\cite{zhang2023llavar} reveals that enhancing the input resolution will significantly increase MLLM's Optical Character Recognition (OCR) ability. 
More and more experiments suggest the presence of an information bottleneck in contemporary vision encoders \cite{tong2024eyes, zhai2023halleswitch}. 
The resolution of the image stands out as a critical factor that constrains the capacity for visual processing.
The study by~\cite{tong2024eyes} highlights that contemporary MLLMs still face systematic challenges, particularly in aligning visual and textual modalities effectively.

There are some works trying to solve the problem. SPHINX \cite{lin2023sphinx}, introduced by Lin et al., employs a multi-layer perception (MLP) to establish connections between visual signals and language. This model leverages multiple vision encoders, amalgamating their output features to construct a robust visual representation. To deal with high resolution image input, SPHINX breaks down input high-resolution images into smaller sub-images and then concatenate visual tokens directly with text tokens. It introduces limitations related to increasing input sequence lengths for the Large Language Model (LLM).

The Monkey model~\cite{li2023monkey} addresses this challenge by incorporating a shared resampler. This approach involves compressing each input subimage using the resampling technique from Flamingo \cite{alayrac2022flamingo} and directly concatenate the visual tokens with text sequence, effectively upscaling the input image resolution to $1344\times896$. However, the reliance on learnable queries within the perceiver architecture for extracting and compressing information from the raw output of the vision encoder raises concerns about the model's adaptability across diverse application scenarios.

\begin{figure}[h]
    \centering
    \includegraphics[width=8cm]{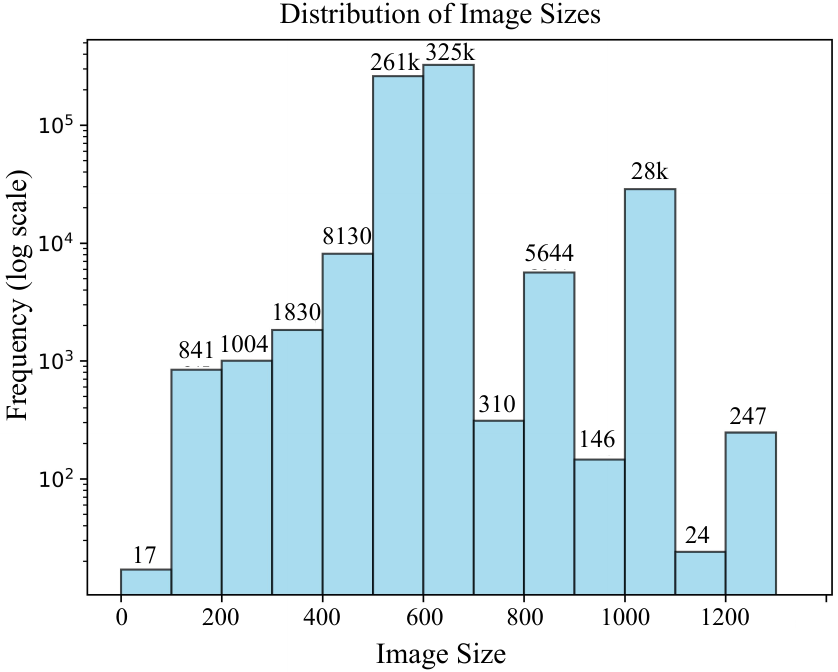}
    \caption{Visualization of the distribution of image sizes from the LLaVA 665k dataset indicates a predominant clustering of resolutions between 500-700, mixed with some high-resolution examples. Dynamic resolution utilization during training is key for efficient resource management.}
    \label{fig:image_size}
\end{figure}

\begin{figure*}[t]
    \centering
    \includegraphics[width=14cm]{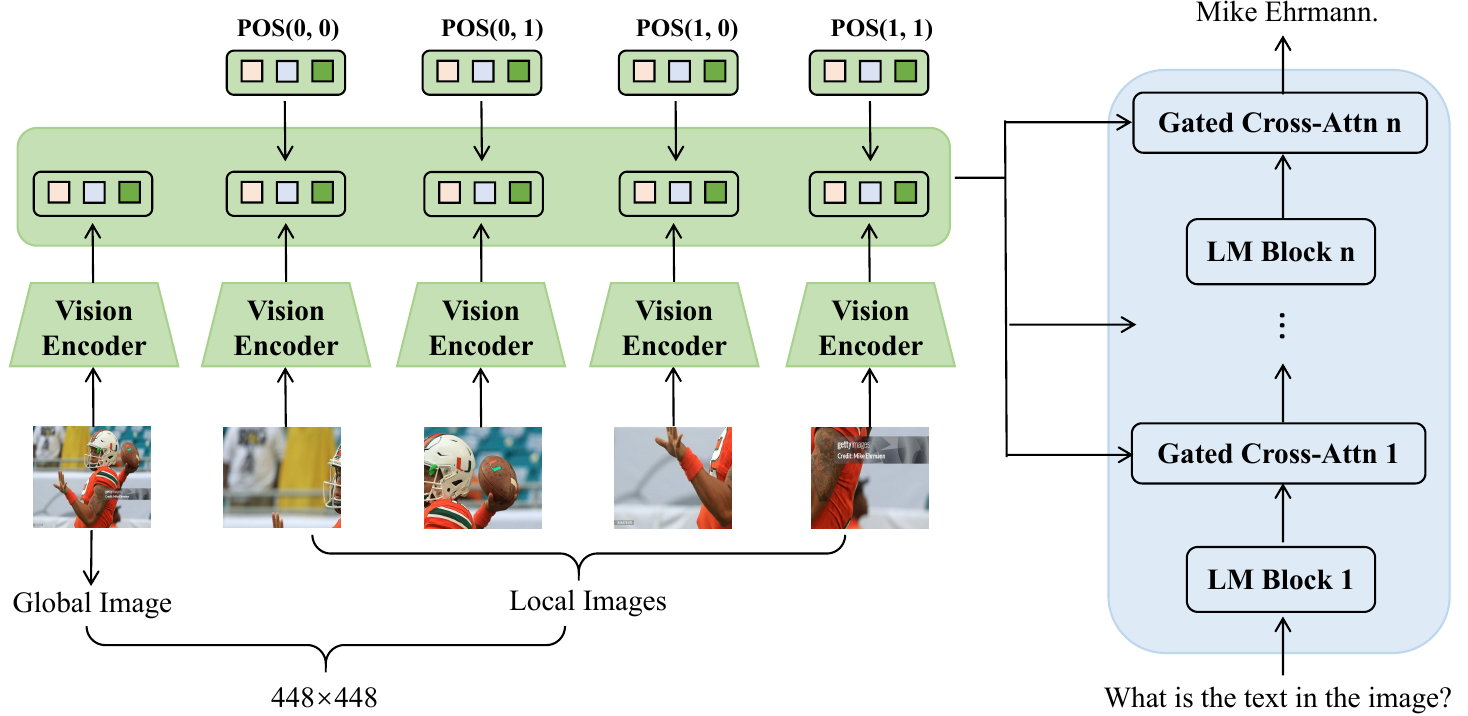}
    \caption{The architectural framework of InfiMM-HD is outlined, where POS(i, j) represents the positional embedding of local patches, with (i, j) indicating their position within the overall image. The model progresses through various training stages, each characterized by selectively training different modules. A detailed explanation of this strategic approach will be elaborated in the following sections.}
    \label{fig:model_arch}
\end{figure*}

\begin{figure*}[t]
    \centering
    \includegraphics[width=16cm]{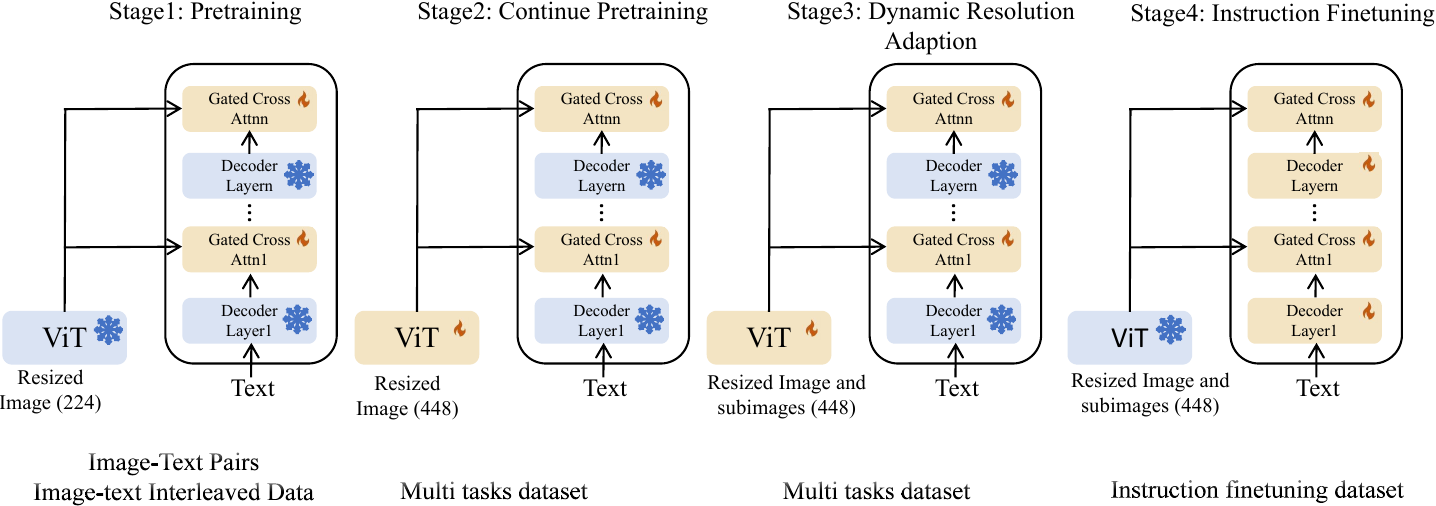}
    \caption{The four stages of InfiMM-HD training pipeline. Each stage is characterized by distinct trainable modules, datasets, and the resolution of images input to ViT. Our experimental findings confirm the efficacy of our approach, demonstrating the benefits of progressively transitioning from 224x224 to high-resolution images.}
    \label{details_pipeline}
\end{figure*}

We assert that the information compression process should intricately intertwine with contextual instructions, allowing for discernment of pertinent details essential for instruction completion. We introduce InfiMM-HD, which establishes connections between vision and language through cross attention mechanism. This departure from the reliance on learnable queries aims to enhance the adaptability and applicability of the model across a broader spectrum of scenarios-more detailed vision perception. Besides, it enables the model to consume high-resolution images at lower cost than previously proposed methods.

\section{Methods}
In this section, we introduce InfiMM architecture and propose a training pipeline for elevating MLLM's input image resolution with reduced cost. To the best of our knowledge, we are the pioneers in achieving HD MLLM using the Flamingo-style architecture.

\subsection{Model Architecture}
\begin{figure}[h]
    
    \centering
    \includegraphics[width=6.5cm]{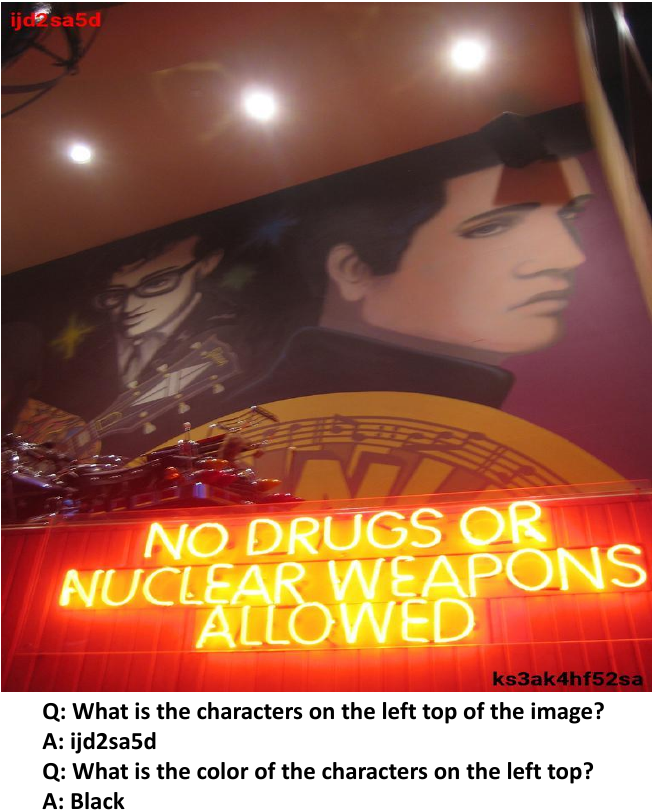}
    \caption{Illustration of data augmentation: Introducing randomly generated characters with diverse colors into arbitrary regions of the image. Corresponding questions are then generated to complement the original query.}
    \label{fig:demo_dataaug}
\end{figure}
The proposed model consists of three components: a Vision Transformer Encoder, a Gated Cross Attention Module, and a Large Language Model. The comprehensive architecture is elucidated in Figure \ref{fig:model_arch}. While the illustration showcases a single image, it is imperative to note that follow flamingo's design, our module can also deal with multiple image as input.
Following prior work \cite{li2023blip2, wang2023cogvlm}, for the Vision Transformer, we employ EVA2-CLIP2-E \cite{sun2023evaclip}, utilizing the output from the penultimate layer as the extracted vision features.
The Gated Cross Attention Module leverages text hidden states as queries, and vision features as keys and values. 
Different from the gating methodology introduced in Flamingo \cite{alayrac2022flamingo}, we incorporate an element-wise $\tanh$ gating mechanism for activation. The language model in this study is instantiated using Vicuna \cite{vicuna2023}.

To ensure an effective assimilation of visual information, the Gated Cross Attention Module is strategically inserted every four layers between the decoder layers of Large Language Model.
This decision stems from our empirical observation that inserting the module every two layers results in approximately 50\% of the gates weights near 0, rendering the cross-attention module ineffective. 
The model showcases an impressive aggregate of 18 billion parameters, intricately allocated among three key components: the Vision Transformer (4.4 billion), the Large Language Model (12.9 billion), and the Gated Cross Attention Module (approximately 0.7 billion).

During this investigation, we depart from the conventional paradigm of the LLaVA-style structure. This deviation is essential due to its compromised compatibility with high-resolution images, as demonstrated in previous studies \cite{li2023monkey, lin2023sphinx}.
Notably, the processing of an image with dimensions $1344\times1344$ yields an extensive token sequence comprising 9217 tokens when employing a patch size of 14. Despite the capability of Large Language Models (LLMs) to accommodate sequence lengths up to 32k, the utilization of 9k tokens per image inherently imposes constraints on the performance of Multimodal Language Models (MLLMs), particularly in scenarios involving multiple images. This limitation is consequential, underscoring the necessity for alternative architectural considerations. These considerations aim to tackle the challenges posed by high-resolution image processing within the context of contemporary language models. We adopt a cross-attention module for the integration of visual information at a reduced dimensionality of 768. This method, in contrast to the LLaVA-style architecture, incurs significantly lower computational costs while accommodating extended sequences. Meanwhile, our experiments demonstrate its effective of extracting visual information.

\begin{table}[h]
\caption{Details on the training data of CPT and DRA.}
\label{data_detail}
\vskip 0.15in
\begin{adjustbox}{width=0.5\textwidth}
\begin{tabular}{c|c|c}
\toprule
Task      & Dataset & Samples \\
\midrule
\multirow{3}{*}{\centering Image Caption}& COCO Caption \cite{chen2015microsoft}& 205k\\
& TextCaps \cite{sidorov2020textcaps}      & 55k\\
& VizWiz Caption \cite{gurari2020captioning}& 55k\\
\midrule
\multirow{5}{*}{\centering General VQA}& VQAV2 \cite{VQA}& 443k\\
& OKVQA \cite{marino2019ok}      & 9k  \\
& VizWiz VQA \cite{gurari2018vizwiz} & 20k \\
& GQA \cite{hudson2019gqa}        & 471k\\
& A-OKQA \cite{schwenk2022okvqa}     & 17k \\
\midrule
\multirow{4}{*}{\centering Text-oriented VQA}& TextVQA \cite{singh2019towards}    & 34k \\
& OCRVQA \cite{mishra2019ocr}     & 166k\\
& STVQA \cite{biten2019scene}      & 26k \\
&DocVQA \cite{mathew2021docvqa}     & 63k \\
&LLaVAR \cite{zhang2023llavar} &16k\\
\midrule
Region Description & VG \cite{krishna2017visual}        & 429k\\
\midrule
Total & -         & 2.00m\\
\bottomrule
\end{tabular}
\end{adjustbox}
\vskip -0.1in
\end{table}
\subsection{Training Details}
We have established a four-stage training procedures for improving MLLMs' capability of processing high-resolution images, as shown in Figure \ref{details_pipeline}.
These stages are denoted as the Pretraining (PT), Continue Pretraining (CPT), Dynamic Resolution Adaption (DRA), and Instruction Finetuning (IFT).

\textbf{Pretraining Stage (PT):} This stage is mainly for initially aligning vision features and language features. During this stage, both the Vision Transformer (ViT) and Large Language Model (LLM) are frozen, only the Gated Cross Attention module is trainable. 
In this stage, all of the images are resized to $224\times224$ to keep the low training cost.

\textbf{Continue Pretraining Stage (CPT):} In this stage, we employ bilinear interpolation of positional embedding to extend the ViT's capability to process image of resolution $448\times448$. 
The ViT and Gated Cross Attention modules are trainable. 
Training datasets mainly focus on image captioning and visual question-answering tasks. 
Detailed information about training datasets is listed in Table \ref{data_detail}. 

\textbf{Dynamic Resolution Adaption (DRA):} In Figure~\ref{fig:image_size}, the outcomes illustrating the sizes of images in the LLaVA-665k dataset \cite{liu2023improvedllava}. Upon meticulous examination of the dataset, it becomes apparent that not all images exhibit resolutions reaching up to 1344. In contrast to conventional practices of uniformly resizing images to a fixed resolution, such an approach incurs unnecessary computational costs, while dynamic image resolution may be cost friendly. To facilitate dynamic resolution inputs, we incorporate the 2D position embedding method proposed in \cite{wang2021translating} for individual sub-images. We adopts dynamic input image resolution, ranging from $448\times448$ to $1344\times1344$, during training. Subsequently, the resized image is divided into sub-images of $448\times448$. We also keep an original image thumbnail of $448\times448$. 
Finally we use ViT to extract features from each sub-image and original image thumbnail, concatenated directly to form the final vision feature. 
We use the same training datasets as CPT stage, and keep both the ViT and Gated Cross Attention modules trainable. 

\textbf{Instruction Finetuning Stage (IFT):} In this final stage, our goal is to make the model better follow user instructions without losing high-resolution visual perception capability.  
Thus, we keep the ViT frozen, but let the Gated Cross Attention modules and LLM trainable. 

The proposed four-stage training pipeline is key to stabilize training while elevate input image resolution gradually.

\section{Experiments}
In this section, we first discuss about experiment setup. Then we show main results of InfiMM-HD and list series of ablation studies to prove the importance of our proposed modules in InfiMM-HD.

\subsection{Setup}

\textbf{Training Dataset.} 
For Pretrainin (PT) stage, training data includes both image-text pairs and interleaved image-text. Image-text pair data includes 140M samples filtered from LAION-2B \cite{li2023blip2}, COYO \cite{kakaobrain2022coyo-700m}, and Laion-coco \cite{laioncoco}. Interleaved image-text data is sampled from MMC4~\cite{zhu2023multimodal} and OBELISIC~\cite{laurencon2023obelics} randomly with $50\%$ for training. 

The datasets utilized in the Continue Pretraining (CPT) and Dynamic Resolution Adaption (DRA) stages are enumerated in Table \ref{data_detail}. 
During the Instruction Finetuning (IFT) stage, we amalgamate datasets from LLaVA-665k \cite{liu2023improvedllava}, LLaVAR \cite{zhang2023llavar}, TextVQA \cite{singh2019towards}, and ScienceQA \cite{lu2022learn}. This fusion is motivated by the fact that the original LLaVA-665k dataset lacks text-oriented samples. Consequently, we supplement this deficiency by incorporating additional data from these diverse sources.

In the IFT stage, we primarily utilize the LLaVA-665k dataset~\cite{liu2023improvedllava} by default. Alongside, we incorporate additional datasets such as TextVQA, LLAVAR, and ScienceQA to enrich our instruction tuning dataset.


\textbf{Text-oriented Data Augmentation.} Due to a scarcity of text-oriented data, we employ a straightforward but effective data augmentation methodology. Concretely, this process involves the incorporation of scene-irrelevant characters into images, followed by the generation of corresponding questions.
In this context, we randomly introduce two sets of characters, each positioned at the left-top and right-bottom corners of the image, respectively, with distinct colors. The questions generated include inquiries such as  ``What character is situated at the left-top of the image?", ``What is the color of the characters located at the right-bottom of the image?", and ``How many characters are present in the left-top region of the image?". It is imperative to note that this data augmentation technique is exclusively applied to the GQA dataset, resulting in the generation of approximately 100,000 question-answer pairs. Remarkably, we observe the effectiveness of this augmentation approach in enhancing our model's proficiency in text recognition. Examples can be found in Figure \ref{fig:demo_dataaug}.

\textbf{Training Details.} 
The training process was facilitated through the utilization of deepspeed \cite{aminabadi2022deepspeed}, and the FusedAdam optimizer was selected to orchestrate optimization tasks. Additional intricacies related to the experimental configurations are thoroughly delineated in the accompanying Appendix \ref{config_detail}.

\begin{table*}[h]
\centering
\caption{Results on general VQA task. The table exclusively presents the performance of our generalist model, showcasing its superiority compared with various models.}
\label{res_text_vqa}
\vskip 0.15in
\begin{adjustbox}{width=1.0\textwidth}
\begin{tabular}{l|l|c|ccccc}
\toprule
Model  & LLM &In-house data & OKVQA & IconVQA & GQA & VQAv2 & ScienceQA\textsubscript{img}\\
\midrule
Flamingo-80B \cite{alayrac2022flamingo}& -         & \checkmark & 50.6  & -       & -   & 56.3  & -      \\
Palm-E-12B   \cite{driess2023palme}  & -         & \checkmark & 60.1  & -       & -   & 77.7  & -      \\
Qwen-VL \cite{bai2023qwenvl}& Qwen-7B   & \checkmark & 58.6  & -       & 59.3& 79.5  & 67.1   \\
Qwen-VL-Chat \cite{bai2023qwenvl}& Qwen-7B   & \checkmark & 56.6  & -       & 57.5& 78.2  & 68.2   \\
CogVLM  \cite{wang2023cogvlm}& Vicuna-7B & \checkmark & 58.9  & -       & -   & \textbf{83.4}  & -        \\
Monkey \cite{li2023monkey}& Qwen-7B   & \checkmark   & 61.3  & -       & 60.7& 80.3  & 69.4      \\
\midrule
BLIP-2 \cite{li2023blip2}& Vicuna-13B& $\times$      & 45.9  & 40.6    & 41.0& -     & -      \\
Shikra \cite{chen2023shikra}& Vicuna-13B& $\times$      & 47.2  & -       & -   & 77.4  & -      \\
mPLUG-Owl2\cite{ye2023mplugowl2}& LLaMA2-7B & $\times$      & 57.7  & -       & 56.1& 79.4  & 68.7   \\
LLaVA 1.5 \cite{liu2023improvedllava}& Vicuna-13B& $\times$      & -     & -       & 63.3& 80.0  & 71.6   \\
Sphinx-2K \cite{lin2023sphinx}& LLaMA2-13B& $\times$      & 62.6  & 50.5    & 63.1& 80.7  & 70.6      \\
\midrule
InfiMM-HD         & Vicuna-13B& $\times$    & \textbf{65.5}  & \textbf{51.3}    & \textbf{63.5}& 82.0  & \textbf{83.6}     \\
\bottomrule
\end{tabular}
\end{adjustbox}
\vskip -0.1in
\end{table*}

\textbf{Evaluation.} We evaluate InfiMM-HD across a diverse array of VQA tasks. For general VQA tasks, we leverage benchmarks such as OKVQA \cite{marino2019ok}, VQAV2 \cite{VQA}, GQA \cite{hudson2019gqa}, and ScienceQA \cite{lu2022learn}. These datasets, while not demanding advanced detail visual perception capabilities from the model, effectively gauge models' ability to understand general scenes and follow user instructions. 
Moreover, to scrutinize our model's fine-grained detail perception capability, we incorporate text-oriented VQA datasets, including TextVQA \cite{singh2019towards}, STVQA \cite{biten2019scene}, and OCRVQA \cite{mishra2019ocr}. 
We assess the logical reasoning capabilities of our model by employing newly introduced benchmarks, including MM-VET \cite{yu2023mmvet}, MME \cite{fu2023mme}, MMbench \cite{liu2023mmbench}, InfiMM-Eval \cite{han2023infimmeval}, and MMMU \cite{yue2023mmmu}. Notably, the MMMU \cite{yue2023mmmu} presents challenging tasks that demand advanced subject knowledge and deliberate reasoning at a collegiate level. These tasks span diverse fields such as physics, chemistry, and biology. 
The MM-VET benchmark assesses the integrated capabilities of models.

\begin{table}[t]
\caption{Evaluation results for text-oriented Visual Question Answering (VQA) task. For STVQA, Monkey randomly samples data from the train set for evaluation.}
\label{res_general_vqa}
\vskip 0.15in
\begin{adjustbox}{width=0.5\textwidth}
\begin{tabular}{l|l|c|ccc}
\toprule
Model  & Res &In-house data  & TextVQA & OCRVQA & STVQA\\
\midrule 
Qwen-VL-Chat\cite{bai2023qwenvl}& $448\times448$ & \checkmark                   & 61.5   & \textbf{70.5}   & -   \\
Monkey \cite{li2023monkey}& $1344\times768$       & \checkmark        & 67.6   & -      & 67.7   \\
\midrule 

UniDoc \cite{gu2022unified}  & $224\times224$ & $\times$  & 40.7  & 34.5   & 30.8\\ 
DocPedia \cite{feng2023docpedia}  & $2560\times2560$   & $\times$              & 60.2  & 57.2   & 45.5\\ 
BLIP-2 \cite{li2023blip2} & $224\times224$   & $\times$                & -      & 40.6   & -   \\
LLaVA1.5 \cite{liu2023improvedllava}  & $336\times336$   & $\times$                & 48.5   & -      & -   \\
Sphinx-2K \cite{lin2023sphinx}  & $768\times768$       & $\times$          & 61.2  & 67.8   & - \\ 

\midrule
InfiMM-HD (all are piexl-only)     & dynamic& $\times$& \textbf{70.7}   & 66.0   & 67.0\\
\bottomrule
\end{tabular}
\end{adjustbox}
\vskip -0.1in
\end{table}

\subsection{Main Results}
We present evaluation results of general VQA and text-oriented VQA tasks in this section. 

Table \ref{res_general_vqa} presents results for general VQA tasks. It is essential to underscore that the scope of evaluations in OKVQA, GQA, and VQAv2 extends beyond the mere assessment of models' visual perception capabilities, they also critically examine the models' ability to utilize prior knowledge effectively, thereby providing a more comprehensive evaluation of models' overall functionality.
Additionally, ScienceQA \cite{lu2022learn}, which comprises 21,000 multimodal multiple-choice questions covering a wide range of scientific subjects, represents a significant expansion of the benchmarking landscape. In these varied tasks, our model has shown remarkable effectiveness, indicating significant performance improvements. By outperforming its closest competitor by an average margin of 3.88\%, our model not only showcases its superior capacity for integrating multimodal information but also its proficiency in leveraging extensive prior knowledge to navigate through a diverse array of questions successfully.

In addition to the general VQA assessment, we further explore our model's detailed visual perception capability by evaluating on text-oriented datasets, including TextVQA, OCRVQA and STVQA, as demonstrated in Figure \ref{fig:demo_textvqa}. Quantitative results are outlined in Table \ref{res_text_vqa}. These results underscore the effectiveness of our proposed high-resolution model in comprehending intricate textual details within images.

We also evaluate InfiMM-HD on recently proposed MLLMs evaluation benchmarks, including MMMU, MM-Vet, InfiMM-Eval, MMB, MME, and POPE. 
Compared with previous VQA datasets, these datasets include more comprehensive evaluation aspects of MLLMs, requiring more complex reasoning capabilities. 
Evaluation results are outlined in Table \ref{res_new_bench}. 
It is noteworthy that no single model excels across all benchmarks, with each model exhibiting its unique strengths and limitations. Our proposed model demonstrates commendable overall performance, highlighting its adaptability and competence across diverse disciplines.

\begin{table*}[ht]
\centering
\caption{Results obtained from benchmarks intricately designed for MLLMs with heightened complexity.}
\label{res_new_bench}
\vskip 0.15in
\begin{adjustbox}{width=1.0\textwidth}
\begin{tabular}{l|ccccccc}
\toprule
Model  & POPE& MME\textsuperscript{P}& MME\textsuperscript{C} & MMB & MM-VET & InfiMM-Eval & MMMU (val) \\
\midrule
BLIP-2 \cite{li2023blip2}& 85.3& 1293.8 & -    & -     & 22.4& -  & - \\
Shikra \cite{chen2023shikra}& -& -      & -    & 58.8  & -& -  & -  \\
LLaVA 1.5 \cite{liu2023improvedllava}& 85.9& \textbf{1531.3}      & 295.4    & 67.7  & 35.4& 32.62  & 36.4  \\
Qwen-VL-Chat \cite{bai2023qwenvl}& -& 1487.5 & \textbf{360.7}  & 60.6 &-   & 37.39& 35.9 \\
Sphinx-2K \cite{lin2023sphinx}& 87.2& 1470.6 & 326.8  & 65.9    & \textbf{40.2}& -  & 32.9      \\
\midrule
Ours         & \textbf{87.9} & 1472.3  & 329.4    & \textbf{71.6}& 38.9  & \textbf{37.42} &\textbf{37.6}  \\
\bottomrule
\end{tabular}
\end{adjustbox}
\vskip -0.1in
\end{table*}

\begin{table}[h]
\centering
\caption{Evaluation Results for models trained with different input resolutions. Here dynamic means the model supports resolution ranging from $448\times448$ to $1344\times1344$. During inference, we don't limit resolution to 1344.}
\label{res_resolution}
\vskip 0.15in
\begin{adjustbox}{width=0.5\textwidth}
\begin{tabular}{l|ccccc}
\toprule
Resolution & GQA & VQAv2 & OCRVQA&DocVQA &  TextVQA  \\
\midrule   
$224\times224$&60.7&78.7&57.6&25.6&50.0\\
$448\times448$&61.3&80.5&58.7&44.9&64.1\\
\midrule
dynamic&63.5&82.0&66.0&55.1&70.7\\
\bottomrule
\end{tabular}
\end{adjustbox}
\vskip -0.1in
\end{table}

\begin{table}[h]
\centering
\caption{Ablation study Results for position embedding. \textbf{w/o PE} means removing the positional embedding.}
\label{pe_impact}
\vskip 0.15in
\begin{adjustbox}{width=0.5\textwidth}
\begin{tabular}{l|ccccc}
\toprule
Resolution & GQA & VQAv2 & OCRVQA&DocVQA &  TextVQA  \\
\midrule   
dynamic (w/o PE)&63.3&81.6&65.4&53.0&70.3\\
dynamic&63.5&82.0&66.0&55.1&70.7\\
\bottomrule
\end{tabular}
\end{adjustbox}
\vskip -0.1in
\end{table}

\begin{table}[h]
\centering
\caption{Evaluation Results for models trained with different resolution. PC here means the model has perceriver resampler.}
\label{ablation_pc}
\vskip 0.15in
\begin{adjustbox}{width=0.5\textwidth}
\begin{tabular}{l|ccccc}
\toprule
Configuration & GQA & VQAv2 &DocVQA &  TextVQA  \\
\midrule   
$224\times224$&60.7&78.7&25.6&50.0\\
$224\times224$ (PC)&57.7&79.0&25.2&48.9\\
\midrule
$448\times448$&61.3&80.5&44.9&64.1\\
$448\times448$ (PC)&56.9&79.5&30.7&56.0\\
\bottomrule
\end{tabular}
\end{adjustbox}
\vskip -0.1in
\end{table}

\subsection{Ablation Study}
To elucidate the direct impact of input image resolution on model performance, we conducted an ablation study. In this investigation, different resolution inputs underwent an identical set of four training stages. 
To ensure experimental equitability, we conduct the same model training without DRA by two epochs on the multi-task datasets.
The results are presented in Table \ref{res_resolution}. An observable pattern suggests that elevating the input images' resolution boosts the efficacy of the model, especially in tasks necessitating an understanding of textual nuances, such as TextVQA \cite{singh2019towards} and DocVQA \cite{mathew2021docvqa}. In contrast, for general VQA tasks, a coarse scene overview is often sufficient for accurate question-answering, underscoring the context-dependent influence of resolution on model efficacy.

Intuitively, as we cropped the input image into subimages, to maintain the spatial information, it is important adding a position embedding for each subimage. To figure out its impact, we carried out ablation study on 2D position embedding, with the results listed in Table \ref{pe_impact}. The findings suggest that removing the position embedding slightly influences model performance. But on DocVQA, it faces apparently degradation. This phenomenon may be attributed to the fact that DocVQA predominantly comprises documents, where the correspondence between various components holds significance, directly reflected through spatial information.

In our model, the perceiver resampler is removed compared with the origin Flamingo. To figure out its impact, we investigated the significance of the perceiver resampler with ablation study. A comparison with models incorporating the perceiver resampler is presented in Table \ref{ablation_pc}. As the table indicates, the perceiver resampler would become an information bottleneck, constraining the model's performance improvement with increased resolution.

\subsection{Limitations}
This study introduces an enhancement to MLLMs to effectively process high-resolution images. Results marks significant advancements in various dimensions. Despite these achievements, certain limitations persist.
In practice, the model exhibits deficiencies in tasks oriented towards text comprehension. 
Our ongoing efforts involve exploring more effective modal alignment strategies while augmenting the dataset to enhance overall model performance.

\section{Conclusions}
In this work, we present InfiMM-HD, an improvement over Flamingo-style MLLM designed for processing high-resolution input images. Our approach leverages a cross-attention mechanism to seamlessly integrate visual information with language model in a low-dimensional space. To address the formidable computational demands associated with high-resolution images, we partition the input high-resolution image into smaller sub-images, each subjected to individual processing using a shared Vision Transformer (ViT) specifically tailored for relatively lower resolutions. Additionally, we establish a four-stage training pipeline to construct the proposed model, ensuring low computational costs are incurred. The proposed model is thus characterized by its comprehensive design and an emphasis on minimizing computational resources.

\section{Broader Impact}
Our model, despite its capabilities, may encounter challenges, including the generation of inaccurate information and susceptibility to perceptual illusions. Furthermore, akin to many machine learning models, it may manifest biases influenced by underlying value systems. Recognizing these potential issues is crucial for ensuring the responsible and ethical deployment of such technologies.


\bibliography{references}
\bibliographystyle{icml2024}

\newpage
\appendix
\onecolumn
\section{Training Configuration}
\begin{figure}[h]
    \label{fig:demo_textvqa}
    \centering
    \includegraphics[width=10cm]{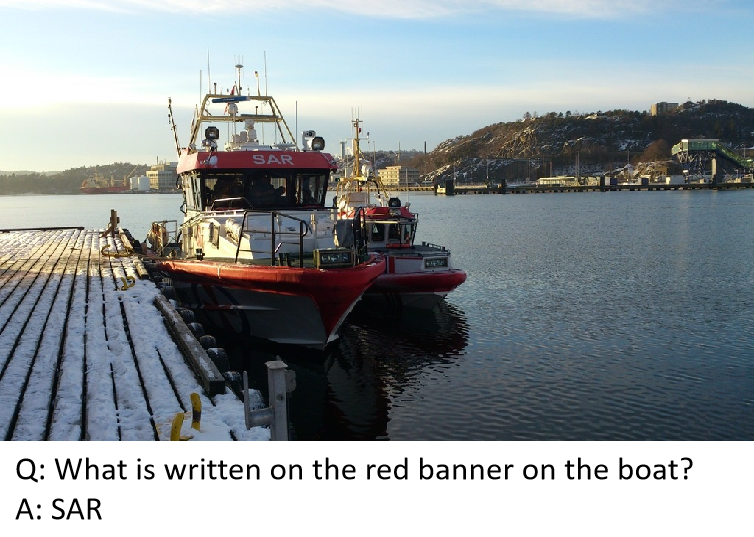}
    \caption{Example of a TextVQA sample, exemplifying its intricate scrutiny of nuanced visual perception.}
\end{figure}

We report the detailed training hyper-parameters and settings of InfiMM-HD in Table \ref{config_detail}. Our model was trained with 64 NVIDIA A100-SXM-80GB. The pretraining stage cost about 72 hours. (If only use the image-text pair, it costs 30 hours.) The remaining three stages cost less than 21 hours.
\begin{table*}[h]
\caption{Details of the training Configuration. For the second stage, we utilize bilinear interpolation to extend the origin ViT to support resolution 448. IT means image text pair. And IIT means interleaved image text sequence.}
\label{config_detail}
\begin{adjustbox}{width=1\textwidth}
\begin{tabular}{ccccc}
\toprule
Configuration & Pretraining& Continue Pretraining  & Dynamic Resolution Adaption & Instruction Finetuning \\
\midrule
ViT init. &  EVA2-CLIP2-E (res 224) &   EVA2-CLIP2-E (res 448)& ViT from 2nd-stage & ViT from 3rd-stage\\
LLM init.&  Vicuna-13b &   Vicuna-13b& Vicuna-13b&Vicuna-13b\\
Gated cross-attention init.&  random & InfiMM-HD 1st stage& InfiMM-HD 2nd stage&InfiMM-HD 3rd stage\\
Image resolution&  224 & 448& dynamic(448-1344)& dynamic(448-1344)\\
ViT sequence length&  257 & 1024& 1025& 1025\\
LLM sequence length&  32 (IT);384 (IIT) & 128& 128& 1024\\
Optimizer&  \multicolumn{4}{c}{AdamW}\\
Optimizer hyperparameter&  $\beta_1=0.9, \beta_2=0.95,eps=1e^{-8}$ & \multicolumn{3}{c}{$\beta_1=0.9, \beta_2=0.999,eps=1e^{-5}$}\\
Peak learning rate&  $1e^{-4}$ &  $1e^{-5}$&  $1e^{-5}$&  $5e^{-6}$\\
Minimum learning rate&  $1e^{-4}$ &  $1e^{-6}$&  $1e^{-6}$&  $5e^{-7}$\\
Learning rate schedule&  \multicolumn{4}{c}{cosine decay}\\
Weight decay&  \multicolumn{4}{c}{0.1}\\
Gradient clip&  \multicolumn{4}{c}{1.0}\\
Training steps&  120k & 8k & 8k & 11k\\
warm steps&6k&400&400&550\\
Global batch size&10240 (IT);768 (IIT)&256&256&64\\
Gradient accumulation steps&2&1&1&2\\
Gradient ACC.&2&1&1&2\\
Numerical precision & \multicolumn{4}{c}{bfloat16}\\
Gradient checkpointing & $\times$ & $\times$ & $\checkmark$&$\times$ \\
Deepspeed Zero Stage & \multicolumn{4}{c}{2}\\
Training resource & \multicolumn{4}{c}{64 NVIDIA A100-SXM-80GB}\\
\midrule
\bottomrule
\end{tabular}
\end{adjustbox}
\end{table*}

\section{Summary of Evaluation Benchmarks}
We provided a detailed summary of evaluation benchmarks we used and their corresponding metrics in Table \ref{eval_detail}. Note that
ANLS means Average Normalized Levenshtein Similarity. 
\begin{table*}[t]
\caption{Details on the test dataset.}
\label{eval_detail}
\begin{adjustbox}{width=1\textwidth}
\begin{tabular}{c|c|c|c|c}
\toprule
Task      & Dataset&Description & Split  & Metric \\
\midrule
\multirow{5}{*}{\centering General VQA}& VQAV2 & VQA on natural images&test-dev&VQA Score($\uparrow$)\\
& OKVQA       & VQA on natural images but need world knowledge &val&VQA Score($\uparrow$)  \\
& IconQA  & Abstract diagram understanding and visual language reasoning&test&EM($\uparrow$) \\
& GQA         & VQA on scene understandig and reasoning&test-dev&EM($\uparrow$)\\
& ScienceQA      & Multimodal multi choice VQA on science filed&test&Accuracy($\uparrow$) \\
\midrule
\multirow{4}{*}{\centering Text-oriented VQA}& TextVQA    & VQA about text in natural scene &val&VQA Score($\uparrow$) \\
& OCRVQA      & VQA on images of book covers&val&EM($\uparrow$)\\
& STVQA       & VQA covering reading and reasoning about text &test&ANLS($\uparrow$)\\
&DocVQA     & VQA on images from documents &test&ANLS($\uparrow$) \\
\midrule
\multirow{6}{*}{\centering Other Benchmarks}& MME     & Evalutaion for MLLM on perception and cognition&Perception and Cognition&Accuracy($\uparrow$)\\
& MM-VET & Dialog style VQA on integrated ability&test&GPT-4 score($\uparrow$) \\
& MMbench  & Comprehensive evalutaion with multi choice VQA&test&Accuracy($\uparrow$) score($\uparrow$) \\
& POPE  & Object hallucination in MLLM&adversial&F1 score($\uparrow$) \\
& InfiMM  & Complex Open-ended Reasoning&test&GPT-4 score($\uparrow$) \\
& MMMU  & College-level multi choice VQA&val&Accuracy($\uparrow$) \\
\bottomrule
\end{tabular}
\end{adjustbox}
\end{table*}


\end{document}